\documentclass{article} %
\usepackage{iclr2027_conference,times}

\usepackage{amsmath,amsfonts,bm}
\usepackage{amsmath,amssymb}

\newcommand{\ques}[1]{$\mathcal{Q}\mathbf{#1}$}

\newcommand{\evalshade}[2]{\begingroup\setlength{\fboxsep}{1pt}\colorbox{#1}{$\mathstrut#2$}\endgroup}
\newcommand{\evalbest}[1]{\evalshade{pink!55}{#1}}
\newcommand{\evalworst}[1]{\evalshade{blue!15}{#1}}

\def\eqref#1{equation~\ref{#1}}

\def\1{\bm{1}}

\DeclareMathAlphabet{\mathsfit}{\encodingdefault}{\sfdefault}{m}{sl}
\SetMathAlphabet{\mathsfit}{bold}{\encodingdefault}{\sfdefault}{bx}{n}

\usepackage{hyperref}
\hypersetup{hidelinks}
\usepackage{url}
\usepackage{booktabs}
\usepackage{tabularx}
\usepackage{longtable}
\usepackage{graphicx}
\usepackage{placeins}
\usepackage{enumitem}
\usepackage{xcolor}

\newcommand{\myparagh}[1]{\noindent\textbf{#1}}

\title{Faster and Better? Benchmark Bugs and \\ Design Limitations Distort the Evaluation \\ of Vision-Language-Action Acceleration}

\author{\hspace{-\tabcolsep}%
\begin{minipage}[t]{\textwidth}
\centering\normalfont
Qiwei Chen\thanks{Equal contribution.\quad $^{\dagger}$Corresponding author: \texttt{gujinyu@sjtu.edu.cn}.}\hspace{0.3em},
Kaijun Zhou\footnotemark[1]\hspace{0.3em},
Nuohui Shi,
Zhiyang Li,
Yuxuan Feng,
and Jinyu Gu\textsuperscript{\ensuremath{\dagger}}\\[5pt]
Shanghai Jiao Tong University
\end{minipage}\hspace{-\tabcolsep}
}

\iclrfinalcopy %
\begin{document}

\maketitle
\fancyhead{} %
\renewcommand{\headrulewidth}{0pt}

\vspace{-2em}
\begin{abstract}
\vspace{-0.6em}
   Simulated manipulation benchmarks are the standard tool for evaluating vision-language-action (VLA) policies
and the acceleration methods that reduce their inference latency for on-robot deployment.
On these benchmarks, we observe that some training-free acceleration methods, which approximate
the baseline policy's computation, achieve higher measured success rates than the baseline itself.
Success rates alone cannot establish whether such gains come from better task execution or from evaluation flaws.
We therefore investigate two kinds of benchmark flaws behind these gains: bugs,
where the implementation does not match the intended task or evaluation protocol, and design limitations,
where success criteria and simulation settings do not fully capture how acceleration affects task execution.
Starting from tasks with anomalous gains, we localize root causes by plotting object trajectories
against checker acceptance regions, and classify the resulting bugs into task consistency,
initialization, and reproducibility.
Extending this audit to seven benchmarks, including RoboTwin, LIBERO-Plus, and VLABench,
we identify 22 bugs of these types and 4 design limitations.
For the latter, we revise permissive success checkers, correct unrealistic object masses,
and add a motion-aware score that favors smoother actions.
Experiments show that bug fixes can reverse method rankings, moving the baseline from last to first on one task.
Addressing design limitations can likewise remove anomalous gains: on another task, the baseline moves
from 21 percentage points behind an accelerated method to 5 points ahead.
Gains attributed to acceleration can therefore be artifacts of the benchmark rather than better task execution.
We release our bug fixes and revised benchmark settings to support trustworthy evaluation of VLA acceleration.

\end{abstract}

\section{Introduction}
\label{sec:introduction}

Vision-language-action (VLA) models~\citep{openvla,smolvla,pi0,pi05,groot}
map visual observations and instructions to robot actions, and acceleration
methods reduce their inference latency for on-robot
deployment~\citep{efficientvla,Efficientsurvey,vlaflash}.
These methods are judged by a simple protocol: report the speedup and show
on standardized manipulation benchmarks~\citep{robotwin2,libero,robodojo,liberopro}
that task success does not drop.
The second half of this protocol rests entirely on the benchmark's success checker.

Training-free acceleration methods~\citep{dpcache,probeflow} approximate the baseline policy's computation
rather than adding capability, yet on some tasks they achieve higher success rates
than the baseline itself.
Such a gain means either that the approximation improves execution or that the
checker scores the altered behavior more generously than the instruction warrants,
and success rates alone cannot tell which.
Therefore, we ask how
benchmark flaws can make accelerated policies appear more successful than they are.

Answering this requires seeing what the robot did, 
not only the checker's verdict.
Recording every rollout adds an 85\% overhead on RoboTwin~\citep{robotwin2}, 
so we instead plot the object and end-effector trajectories recorded during evaluation against the checker's acceptance region
and inspect videos only where this view flags a problem.
This inspection reveals two patterns.
In a \emph{false baseline failure}, 
the baseline completes the instructed task but is rejected: 
in \emph{Place Shoe}, the checker demands a toe orientation the
prompt never mentions, failing the baseline while passing the accelerated variant that reproduces the demonstrations' orientation (Figure~\ref{fig:shoe-failure}).
In a \emph{false acceleration success}, the accelerated policy is accepted without completing the task: in \emph{Put Bottles Dustbin}, it never grasps the bottle but
knocks it off the table, the bottle rolls into the acceptance region, and the checker counts a success (Figure~\ref{fig:dustbin-success}).

Tracing such cases across tasks and benchmarks, we find two kinds of flaws behind them.
\emph{Benchmark bugs} are places where the implementation departs from the intended
task or evaluation protocol.
They fall into three categories: instructions inconsistent with checkers and
demonstrations, invalid initialization of prompts or scenes, and evaluation
conditions that cannot be reproduced.
Each corrupts the comparison differently, by scoring rollouts against unstated
requirements, by changing the task before the policy acts, or by evaluating
baseline and accelerated policies under different conditions.

\emph{Design limitations} arise where the benchmark works as implemented but does not capture how acceleration changes execution.
Checkers can be permissive: they may accept an object entering a target region
without verifying the manipulation that put it there, and their distance
thresholds make success rates sensitive to hyperparameters the instruction
never exposes.
Simulation can be idealized: unrealistic object masses let the same actions move
an object differently in simulation and reality, and action smoothing that
differs from real robots hides the jitter that acceleration adds to policy outputs.

Across seven benchmarks, this audit yields 22 bugs and 4 design limitations.
We address them and re-evaluate the baseline against three accelerated variants.
The corrections change conclusions, not only numbers.
On LIBERO-PRO~\citep{liberopro}, bug fixes raise the baseline's success rate on one task by 52.8 percentage
points and move it from last to first; on RoboTwin, correcting object masses
moves the baseline from 21 percentage points behind an accelerated variant to 5 points ahead.

Our contributions are threefold:
\begin{itemize}[topsep=2pt,itemsep=1pt,parsep=0pt,partopsep=0pt]
    \item We characterize misleading acceleration gains and introduce trajectory and acceptance-region visualization
    to provide evidence beyond success rates.
    \item We uncover 22 benchmark bugs and 4 design limitations.
    \item We re-evaluate acceleration after bug fixes and targeted design revisions,
    show that method rankings can reverse, and distill practical guidelines
    (Appendix~\ref{app:practical-guidelines}).
\end{itemize}

\vspace{-1em}

\section{Related Work}

\vspace{-0.5em}
\myparagh{Robotic Manipulation Benchmarks.}
Robotic manipulation benchmarks evaluate policies across diverse tasks and
environments~\citep{rlbench,calvin}. LIBERO covers spatial, object, and goal variations as well as
long-horizon tasks~\citep{libero}, while LIBERO-PRO and LIBERO-Plus extend
its evaluation with task and environment perturbations to examine
robustness~\citep{liberopro,liberoplus}.
Meta-World~\citep{metaworld} and RoboTwin~2.0~\citep{robotwin2} cover multi-task and bimanual manipulation,
while RoboCasa targets kitchen activities~\citep{robocasa} and VLABench emphasizes commonsense and
long-horizon reasoning~\citep{vlabench}.
SIMPLER reduces visual and control gaps in simulation~\citep{simpler},
and RoboDojo unifies simulation and real-world evaluation~\citep{robodojo}.
These benchmarks commonly measure performance by success rates from binary
task-success judgments, while RoboDojo additionally reports a graded score of
partial task progress.

\myparagh{Training-free VLA Inference Acceleration.}
Training-free acceleration methods improve inference efficiency by mitigating
computational redundancy. 
EfficientVLA combines language-model layer pruning,
task-aware visual token selection, and intermediate-feature caching 
to reduce computation across multiple components of the
VLA model~\citep{efficientvla}. ProbeFlow and DP-Cache reduce inference
latency by trimming redundant computations within the denoising phase of VLA
models, whereas \emph{Knowing When to Stop} lowers the overall inference
frequency by adaptively tuning the number of actions executed per action chunk
\citep{probeflow,dpcache,knowingwhentostop}. 
Typically evaluated against native baselines, these approaches use task success
rate as a central measure of task performance. Such comparisons do not by
themselves establish whether higher success rates reflect better execution
of the task instructions.

\myparagh{Reliability of Robotic Manipulation Evaluation.}
RoboEval supplements binary success with measures of task progress, 
execution efficiency, bimanual coordination, and safety, 
revealing behavioral differences between policies with similar success rates~\citep{roboeval}. 
\citet{whatwebenchmark2026} audit five manipulation benchmarks and identify four threats to benchmark validity: 
shortcut solvability, insufficient evidence of statistical significance, 
creeping overfitting, and dependence on the source of training demonstrations. 
These issues weaken the connection between benchmark scores and manipulation capabilities. 
These works primarily examine evaluation reliability through
behavioral measurement, task and test-distribution design, statistical analysis, and control of training data. 
We examine how benchmark corrections and evaluation refinements change comparisons between baseline and accelerated policies.
\vspace{-0.8em}
\section{Characterizing the Evaluation of VLA Acceleration}
\label{sec:problem-definition}
\label{sec:acceleration-evaluation}
\vspace{-0.8em}

This section introduces the components of VLA benchmarks and examines how
instructions and demonstrations shape the behaviors assessed by success-rate
metrics. We then use representative cases to illustrate their implications
for evaluating VLA acceleration.

\vspace{-0.8em}
\subsection{Benchmark Components}
A simulated manipulation benchmark comprises simulation environment, task specifications, initialization procedures, success checkers,
instruction construction, expert training demonstrations, and evaluation metrics.
Table~\ref{tab:benchmark-construction} summarizes instruction construction,
training demonstration sources, and evaluation metrics in representative
benchmarks. Across these simulation benchmarks, success is evaluated using
rule-based checks on simulator states, with task-specific combinations of
conditions, targets, and hyperparameter thresholds.

\begin{table}[!htbp]
\centering
\vspace{-1.5em}
\caption{Benchmark components. Instruction, Demos, and Evaluation denote
instruction construction, training demonstration sources, and the success signal, respectively.}
\label{tab:benchmark-construction}
\label{tab:benchmark-eval-units}
\footnotesize
\setlength{\tabcolsep}{3pt}
\renewcommand{\arraystretch}{1.0}
\resizebox{\linewidth}{!}{%
\begin{tabular}{@{}cccc@{}}
\toprule
Benchmark & Instruction & Demos & Evaluation \\
\midrule
LIBERO~\citep{libero}
& Not specified
& Human teleoperation
& Binary \\
\addlinespace[2pt]
LIBERO-PRO~\citep{liberopro}
& Based on LIBERO
& Reused from LIBERO
& Binary \\
\addlinespace[2pt]
LIBERO-Plus~\citep{liberoplus}
& LLM-rewritten
& Reused from LIBERO
& Binary \\
\addlinespace[2pt]
RoboTwin~\citep{robotwin2}
& LLM-generated
& Scripted experts, motion planning
& Binary \\
\addlinespace[2pt]
RoboCasa~\citep{robocasa}
& Not specified
& Human teleoperation, MimicGen
& Binary \\
\addlinespace[2pt]
VLABench~\citep{vlabench}
& LLM-rewritten
& Scripted experts, motion planning
& Binary, stage scores \\
\addlinespace[2pt]
RoboDojo~\citep{robodojo}
& Not specified
& Human teleoperation, motion planning
& Binary, stage scores \\
\bottomrule
\end{tabular}%
}
\end{table}

The simulation environment governs physics and rendering, the task specification defines the intended behavior.
Initialization procedures define the
starting state, scenario, physical parameters, and control settings.
Instructions and demonstrations influence the behavior being scored: 
the instruction specifies the requested task, and training demonstrations provide expert examples
of its execution.
Except for LIBERO variants, these benchmarks generate instructions by filling
template slots with task-specific information. 
The evaluation metric records how often the resulting behavior
satisfies the checker. 
The benchmark components provide the context for interpreting
success-rate gains from acceleration.

\vspace{-0.6em}
\subsection{How Benchmark Flaws Distort Acceleration Comparisons}
\vspace{-0.4em}
Training-free acceleration studies evaluate their methods on several
manipulation benchmarks and report success-rate increases in some settings.
ProbeFlow~\citep{probeflow} evaluates on MetaWorld and LIBERO, with a small
numerical increase on MetaWorld.
DP-Cache~\citep{dpcache} evaluates on LIBERO and reports gains on its Spatial
and Long suites.
However, these studies do not report whether the additional rollouts counted
as successful satisfy the task instructions beyond passing the benchmark
success checkers.

Motivated by these reported gains, we examine whether benchmark flaws can
change the measured performance of baseline and accelerated policies on RoboTwin. 
We inspect rollouts alongside the task instructions, available
demonstrations, and checker outcomes. The following cases illustrate how
checkers can reject valid task execution or accept behavior that violates
the task instructions.
Configurations of the acceleration methods are detailed in
Section~\ref{sec:experimental-setup}.

\begin{figure}[!htbp]
\centering
\setlength{\abovecaptionskip}{3pt}
\includegraphics[width=\linewidth]{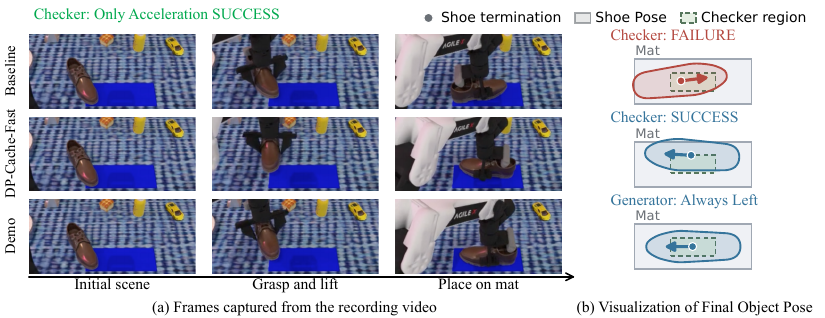}
\caption{Task \emph{Place Shoe}. Simplified prompt: ``Pick up the footwear from the table and place it on the mat.''
(a)~Baseline and DP-Cache-Fast rollouts alongside a rule-based expert demonstration.
(b)~The final poses show that both shoes lie on the mat.}
\label{fig:shoe-failure}
\vspace{-1.2em}
\end{figure}

\begin{figure}[t]
\centering
\setlength{\abovecaptionskip}{3pt}
\includegraphics[width=\linewidth]{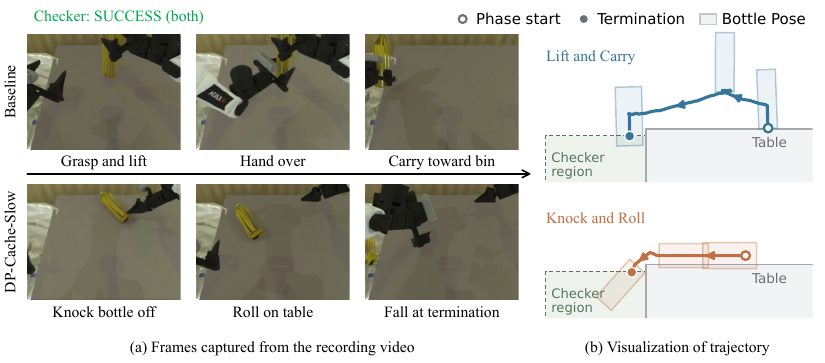}
\caption{Task \emph{Put Bottles Dustbin}. Simplified prompt: ``Pick the bottle with
two distinct parts, drop it into the garbage bin.''
(a)~The recording rollout videos.
(b)~The trajectory visualizations show how both bottles enter the checker's
acceptance region, serving as an alternative to rollout videos.}
\label{fig:dustbin-success}
\vspace{-1.6em}
\end{figure}

\vspace{-0.5em}
\subsubsection{A Benchmark Bug Causes a False Baseline Failure}
\vspace{-0.5em}

We evaluate acceleration methods on the \emph{Place Shoe} task.
In the example in Figure~\ref{fig:shoe-failure}, starting from the same initial state, 
the baseline places the shoe facing right, while DP-Cache-Fast places it facing left. 
Both placements satisfy the prompt, but the checker rejects the baseline and accepts DP-Cache-Fast, 
enforcing a toe orientation that the prompt does not specify.
DP-Cache-Fast's left-facing placement is learned from the rule-based training
demonstrations. The checker follows the same orientation requirement as the
demonstrations, causing it to reject the baseline's otherwise valid placement.
After the benchmark fix, which also accepts right-facing placements, the baseline
success rate rises from $65\%$ to $74\%$, while DP-Cache-Fast rises from $68\%$ to $69\%$.
The baseline thus shifts from trailing DP-Cache-Fast by 3 percentage points to leading it by 5 points.

\vspace{-0.5em}
\subsubsection{A Design Limitation Allows a False Acceleration Success}
\label{sec:characterize-limitation}
Figure~\ref{fig:dustbin-success} illustrates the \emph{Put Bottles Dustbin}
task with baseline and DP-Cache-Slow rollouts from the same initial scene.
Although rolling the bottle off the table does not
obey the instruction to pick it up, the rollout is still classified as successful.
The checker tests whether the bottle enters the acceptance region,
without verifying the instructed manipulation process.
This case exposes a limitation of the success criterion: a higher reported score may not necessarily
indicate better execution of the instructed manipulation.
After adding a check that requires the gripper to be positioned above
the bin within a time window before the bottle falls, the baseline success
rate remains at $24\%$, whereas DP-Cache-Slow drops from $28\%$ to $22\%$.
The baseline thus shifts from trailing DP-Cache-Slow by 4 percentage points to leading it by 2 points.

These cases show how benchmark flaws produce anomalous gains by
rejecting valid baseline behavior or accepting accelerated behavior
that violates task instructions.
Building on these observations, we investigate benchmark bugs and design
limitations in Sections~\ref{sec:benchmark-bugs} and~\ref{sec:design-limitations}, respectively.

\vspace{-0.8em}
\section{Workflow for Identifying Benchmark Flaws}

\vspace{-0.4em}
To investigate the anomalous gains observed above, we use the following
workflow to identify benchmark flaws.

\myparagh{Select tasks through anomalous gains.}
Success-rate gains motivate inspection but do not establish a flaw.
First verify that baseline and accelerated runs use the same task instance
and evaluation settings, investigating any initialization or reproducibility
mismatch. For matched runs with different success signals, inspect both
rollouts for false rejection or acceptance relative to the instruction.

\myparagh{Localize the cause through lightweight visualization.}
Plot logged object and end-effector trajectories alongside gripper states
and checker acceptance regions, marking success-trigger times.
Compare these views with instructions and demonstrations, inspect the
relevant code, and have domain experts manually verify suspected flaws.
This view supports inspection without rendering and encoding every rollout,
avoiding approximately 85\% video-recording overhead
(160.4 seconds/episode with recording vs.\ 86.4 without).
Appendix~\ref{app:recording-overhead} reports detailed timing measurements.

\myparagh{Extend the audit by failure mechanism.}
Inspect tasks sharing the affected templates, loaders, checkers, or scene
utilities, and components with equivalent roles in other benchmarks.
Verify each case against its task specification and implementation,
including tasks without anomalous gains.

By using acceleration methods to uncover benchmark flaws, we aim to enable
fairer comparisons among variants of the same embodied intelligence model policy.

\vspace{-1em}
\section{Localizing Benchmark Bugs}
\label{sec:benchmark-bugs}

\vspace{-0.5em}
Following the workflow above, we identify 22 bugs across the 7 benchmarks
in Table~\ref{tab:benchmark-construction}. These bugs fall into 3 categories:
task consistency, initialization, and reproducibility.
Section~\ref{sec:bug-fixing-results} compares evaluation results before and after bug fixes.

\begin{figure}[!htb]
\vspace{-1em}
\centering
\setlength{\abovecaptionskip}{2pt}
\includegraphics[width=\linewidth,trim=0 0 0 20bp,clip]{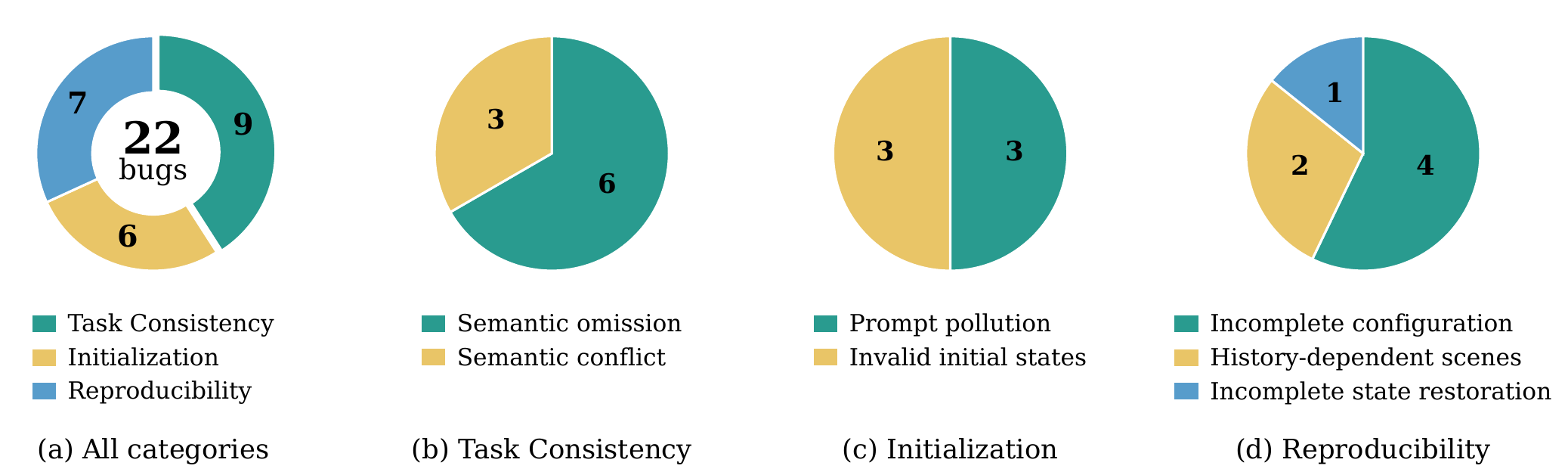}
\caption{Distribution of the 22 benchmark bugs.
(a) Counts across all categories; (b)--(d) subcategory counts within task
consistency, initialization, and reproducibility, respectively.}
\label{fig:benchmark-bug-counts}
\vspace{-1.4em}
\end{figure}

\subsection{Task Consistency: Instruction--Checker--Demo Misalignment}
\label{sec:task-consistency}

\vspace{-0.4em}
As Table~\ref{tab:benchmark-construction} shows, 
some benchmarks use LLM to expand task instructions, increasing linguistic diversity for evaluating generalization. 
However, diverse wording does not guarantee preservation of task semantics, 
and combining language templates with scenario information can introduce unexpected conflicts.
Figure~\ref{fig:benchmark-bug-counts}(b) summarizes these inconsistencies.

\myparagh{(1) Semantic omission.}
Some instructions omit constraints retained in the demonstrations and success checkers.
As shown earlier, \emph{Place Shoe}'s checker restricts toe orientation beyond
the instruction's placement requirement. Relaxing this orientation constraint
aligns the checker with the semantics of the delivered prompt.
In RoboTwin's \emph{Handover Block} task, the prompt is “Grab the red block with the left arm.”
It omits the handover and placement behavior shown in the demonstrations:
the left arm passes the block to the right arm, which places and releases it on the pad.
The checker requires placement on the pad and an open right gripper, rejecting grasp-only completion.
Similar omissions occur in RoboTwin's \emph{Blocks Ranking Size}, \emph{Rotate QRCode} and \emph{Adjust Bottle},
and VLABench's \emph{Select Mahjong}.

\myparagh{(2) Semantic conflict.}
Instructions, checkers, and demonstrations specify different operations.
In RoboTwin's \emph{Place Bread Basket} task, some instructions require two-arm
grasping of adjacent bread, but the corresponding demonstrations use one arm.
The checker checks final bread positions without verifying arm use, allowing one-arm execution to pass.
In RoboTwin's \emph{Place Bread Skillet}, some instructions request left-arm
bread transport, but the corresponding demonstrations use the right arm,
and the checker does not verify which arm is used.
In VLABench's \emph{Select Painting}, the instruction asks the robot to
grasp a painting, but the demonstration presses a button and passes the checker.

These cases show that instructions, demonstrations, and checkers do not
always specify the same task requirements. As a result, the success signal may
not accurately reflect whether a policy follows the given instruction.

\vspace{-1em}
\subsection{Initialization: Prompt and Initial-State Errors}
\label{sec:initialization-bugs}
\vspace{-0.4em}

Section~\ref{sec:task-consistency} concerns instructions that misstate or omit
task requirements. Here, the error occurs during initialization: the loader
selects the wrong task description or adds unrelated text, or pairs a scene
with incompatible object states, as shown in Figure~\ref{fig:benchmark-bug-counts}(c).

\myparagh{(1) Prompt pollution.}
The received instruction does not match the task.
In LIBERO-PRO's \emph{Turn On The Stove} task, the policy receives
the ``turn on the stove'' prompt, but the checker requires it off. 
The loader derives the prompt from an unchanged filename instead of the updated task definition. 
LIBERO-Plus appends task-specific metadata as instruction suffixes
or mistakes prompts written for LLM-based rewriting for task instructions.

\myparagh{(2) Invalid initial states.}
The initial state violates intended scene conditions.
In a LIBERO-PRO \emph{Environment Perturbation} task,
mismatched generation and loading directories cause a new scene to load an old
state file without validating object positions. When a living-room tabletop
scene is replaced with a kitchen scene, objects retain their original positions,
fall from midair, and bounce off the table, preventing task completion.
In RoboCasa, source and destination counters can resolve to the same physical object, 
making transport successful at initialization. RoboDojo's \emph{Organize Table}
requests placing remaining objects in a drawer, but no such objects exist on the tabletop.

These defects alter the task presented to the policy or its initial conditions,
potentially making a task impossible to complete or already successful before
execution. The resulting scores can therefore distort comparisons between
baseline and accelerated policies on the intended tasks.

\vspace{-0.6em}
\subsection{Reproducibility: Uncontrolled Randomness and Incomplete Replay}
\label{sec:reproducibility-bugs}
\vspace{-0.4em}

To isolate the effects of acceleration, comparisons must hold non-policy variables fixed.
We identify 3 types of implementation defects that compromise
this control, as summarized in Figure~\ref{fig:benchmark-bug-counts}(d).

\myparagh{(1) Incomplete evaluation configuration.}
Identical seeds can yield different evaluation conditions.
In RoboTwin, the same scene can receive different instructions because the
episode seed does not control instruction sampling. VLABench similarly fails
to pass the requested seed to its independent scene generator. LIBERO-Plus
samples non-default task orders before setting the seed, allowing identical
configurations to produce different task sequences.
LIBERO leaves MuJoCo unpinned, allowing version changes to alter rendered observations.

\myparagh{(2) History-dependent scene construction.}
Matching trial indices need not identify the same scene.
In RoboCasa, the environment generates a scene for each trial. Even when two
runs start with the same seed, they can generate different scenes for the same trial. 
In RoboTwin, the same seed can produce different clutter positions depending
on how many episodes have already run in the process.
Matching trial indices thus does not ensure matched scenes.

\myparagh{(3) Incomplete state restoration.}
In some RoboCasa tasks, reloading a saved successful state can produce a failure judgment. 
The robot and objects return to their saved positions, but the checker
still uses target positions from a different scene. Restoring the visible scene
therefore does not restore the conditions used to judge success.

These problems break consistency in inputs, scenes, or scoring conditions.
Identical seeds, trial indices, or even saved states are therefore insufficient
to ensure matched evaluation conditions. Measured success-rate differences can
include effects of non-policy variables, making them difficult to attribute
specifically to acceleration methods.

\vspace{-1em}
\section{Benchmark Design Limitations}
\label{sec:design-limitations}

\vspace{-0.7em}
Beyond implementation bugs, we examine two categories of benchmark design
limitations: permissive success checkers and simulation-to-reality gaps.
Section~\ref{sec:limitation-results} evaluates how addressing these
limitations affects policy comparisons.

\begin{figure}[!htb]
\vspace{-6pt}
\centering
\setlength{\abovecaptionskip}{4pt}
\begin{minipage}[t]{0.32\linewidth}
\centering
\includegraphics[width=\linewidth]{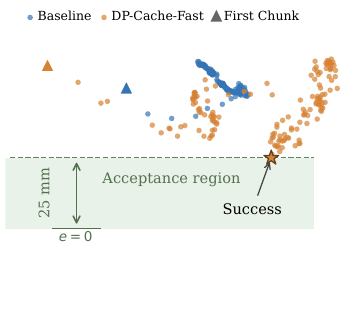}
\vspace{-37pt}\par\small (a) Spatial distributions
\end{minipage}%
\begin{minipage}[t]{0.32\linewidth}
\centering
\includegraphics[width=\linewidth]{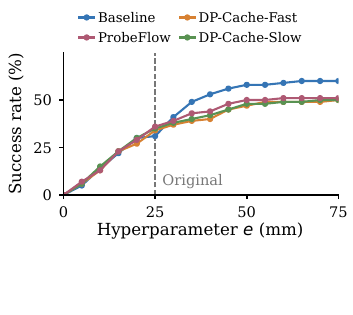}
\vspace{-37pt}\par\small (b) Threshold sensitivity
\end{minipage}%
\begin{minipage}[t]{0.36\linewidth}
\centering
\includegraphics[width=\linewidth]{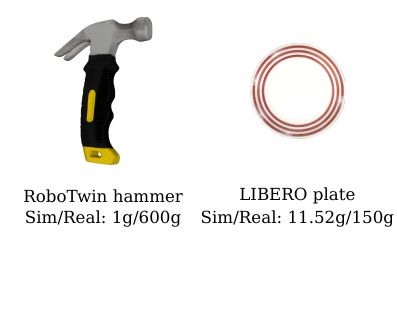}
\vspace{-37pt}\par\small (c) Object masses
\end{minipage}
\par\vspace{-16pt}
\caption{Benchmark design limitations arising from fixed acceptance
thresholds and unrealistic object masses.
(a,b) Spatial distributions of \emph{Scan Object} samples relative
to the acceptance region and success rates under different acceptance
thresholds, respectively.
(c) Discrepancies between simulated and real-world object masses.}
\label{fig:design-limitations}
\vspace{-1em}
\end{figure}

\subsection{Permissive Checkers}
\label{sec:permissive-checkers}

\myparagh{(1) Incomplete checker criteria.}
Current checkers incompletely capture the execution process, allowing
success without verifying how the goal was reached, as detailed in Figure~\ref{fig:dustbin-success}.
In VLABench's \emph{Select Drink} task, the checker only requires
the target drink to leave the container, so a juice carton dropped
during extraction still counts as a success once it leaves
the fridge region.

\myparagh{(2) Reliance on fixed acceptance thresholds.}
A single acceptance threshold may bias policy comparisons.
In \emph{Scan Object}, distance thresholds define the acceptance region
used to determine success.
As shown in Figure~\ref{fig:design-limitations}(a), which plots the spatial
distributions from one rollout per policy,
DP-Cache-Fast exhibits variances in axial distance and transverse error $e$ that are $11.8\times$ and $4.5\times$ those of
the baseline, respectively.
Given different spatial distributions, success rates
may depend on acceptance thresholds.
To assess whether the policy comparison is robust to threshold choice,
we sample the parameters defining the acceptance region and re-evaluate across these settings
(Figure~\ref{fig:design-limitations}(b)).

\vspace{-0.5em}
\subsection{Simulation-to-Reality Gaps in Physical Execution}
\label{sec:physical-execution-gaps}
\vspace{-0.4em}

Beyond visual variation addressed by domain
randomization~\citep{domainrandomization}, simulation fidelity depends on
action controllers and object properties.

\myparagh{(1) Unrealistic object masses.}
\label{sec:simulation-parameters}
In RoboTwin, over 60\% of the audited dynamic object creation sites
assign the default mass of 10\,g, while even individually specified
masses can be unrealistic: the hammer in \emph{Beat Block Hammer}
is assigned only 1\,g, compared with its real-world mass of 600\,g.
RoboCasa and LIBERO compute masses for most objects from collision
geometry using a density of 100\,kg/m$^3$, which may not reflect
their actual materials.
For example, a plate in LIBERO is assigned a mass of just 11.52\,g,
compared with its real-world mass of 150\,g.
These discrepancies illustrate how visually realistic objects can
have implausible physical properties, as shown in Figure~\ref{fig:design-limitations}(c).
Such mass settings can alter object responses to the same policy
actions and contribute to the simulation-to-reality gap~\citep{dynamicsrandomization};
Table~\ref{tab:mass-results} reports the evaluation results.

\begin{figure}[!htbp]
\centering
\vspace{-0.6em}
\includegraphics[width=\linewidth]{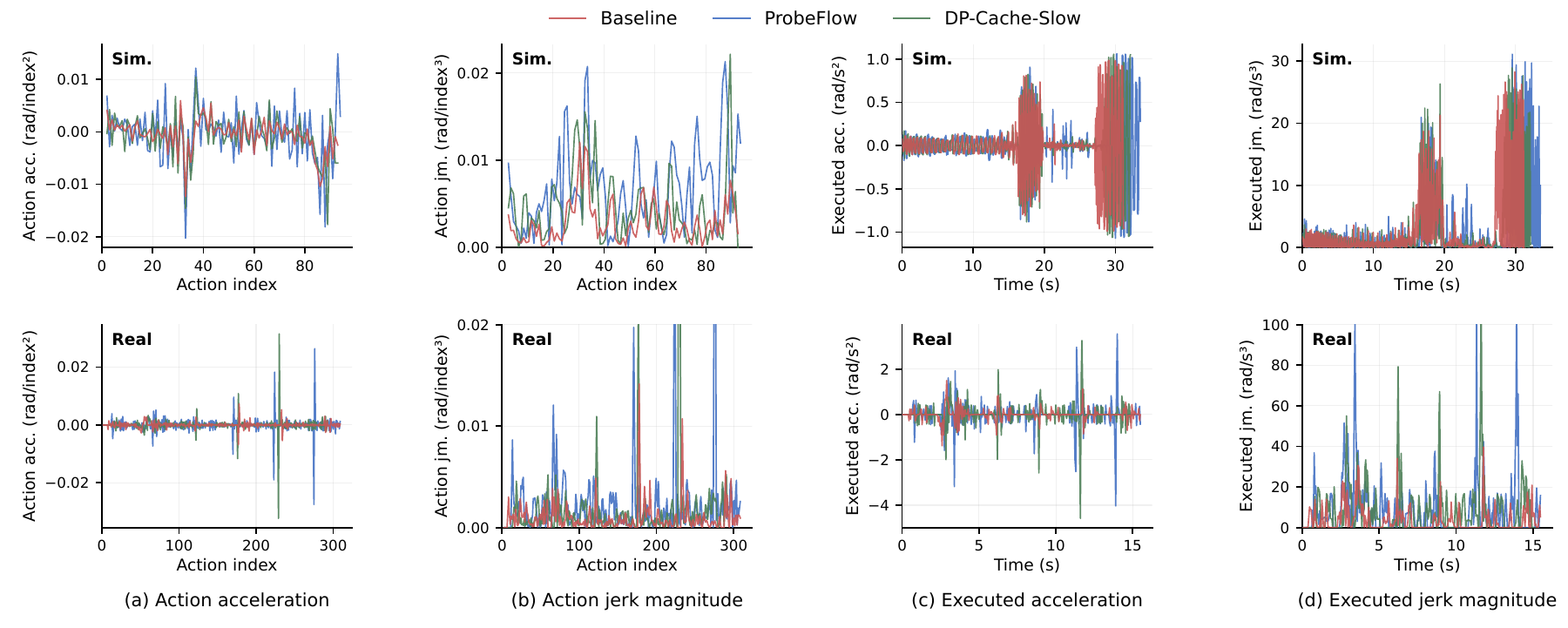}
\addtolength{\abovecaptionskip}{-19pt}
\caption{Policy-action fluctuations and executed motion in simulation (top)
and on the real robot (bottom). Acceleration and jerk magnitude are
abbreviated as acc. and jm., respectively.}
\label{fig:action-execution-comparison}
\vspace{-1.4em}
\end{figure}

\vspace{-0.3em}
\myparagh{(2) Differences in action processing.}
\label{sec:action-execution}
We compare policy outputs and executed motion on \emph{Pick Dual Bottles}
in simulation and on a real robot
(Figure~\ref{fig:action-execution-comparison}).
Following prior smoothness analyses~\citep{vla-rail, roboeval}, we measure
executed acceleration (rad/s$^2$) and executed jerk magnitude (rad/s$^3$).
For denormalized outputs, action acceleration (rad/index$^2$) and
action jerk magnitude (rad/index$^3$) are computed over action indices
rather than time, before post-processing.

In simulation, DP-Cache-Slow and ProbeFlow have mean action jerk
magnitudes of $1.45\times$ and $2.16\times$ the baseline's,
respectively, while their mean executed jerk magnitudes are only
$0.94\times$ and $0.92\times$ the corresponding baseline value.
On the real robot, the latter ratios rise to $1.35\times$ and $1.43\times$.
Thus, the smoothness advantage in simulation does not
necessarily transfer to real execution.

In simulation, time-optimal path parameterization (TOPP)~\citep{topp}
retimes paths under motion constraints, changing execution duration
and potentially the measured jerk.
Control-frequency limits and differences in low-level motor-control
implementations may explain the discrepancy between simulation
and real execution.
We therefore use action smoothness before post-processing as a
complementary ranking criterion for on-robot deployment.
For each paired episode, we rank all four methods by action variation
in ascending order and assign rank-based scores, with failed rollouts
receiving zero points. Appendix~\ref{app:motion-ranking} provides the details.
This score favors smoother policy outputs by accounting for
action fluctuations before post-processing.

\vspace{-1em}
\section{A Trustworthy Evaluation of VLA Acceleration}
\label{sec:trustworthy-evaluation}

\vspace{-0.5em}
Our evaluation addresses two questions:
\par
\noindent\ques{1}: How do benchmark bug fixes affect the measured gains of VLA acceleration methods? (\S~\ref{sec:bug-fixing-results})
\par
\vspace{-0.5em}
\noindent\ques{2}: How do checker design and physical parameters shape the interpretation of these gains? (\S~\ref{sec:limitation-results})

\vspace{-0.5em}
\subsection{Experimental Setup}
\label{sec:experimental-setup}

\vspace{-0.5em}
\myparagh{Hardware and models.}
Our evaluation infrastructure uses NVIDIA A800 and RTX 4090 GPUs. 
We build on public benchmark implementations and community-released $\pi_{0.5}$~\citep{pi05} checkpoints. 
Further hardware and software details are provided in Appendix~\ref{app:hardware-models}.

\myparagh{Benchmarks and Acceleration methods.}
We audit seven benchmarks (Section~\ref{sec:benchmark-bugs}) and evaluate the tasks listed below. 
We compare Baseline with DP-Cache-Fast ($S=5$), DP-Cache-Slow ($S=2$)~\citep{dpcache}, and
ProbeFlow ($\epsilon=0.008$)~\citep{probeflow}, which achieve speedups of
$2.56\times$, $1.62\times$, and $1.49\times$, respectively.
Configuration details are shown in Appendix~\ref{app:method-configurations}.

\myparagh{Metrics.}
We report success rates (SR) and gains over the baseline before and after revision.
For motion-aware scoring, revised values report graded scores rather than success rates.
Measurement details are provided in Appendix~\ref{app:measurement-details}.
Additional results are shown in Appendix~\ref{app:additional-results}.

\vspace{-0.8em}
\subsection{Bug Fixing Results}
\label{sec:bug-fixing-results}
\vspace{-0.4em}

Table~\ref{tab:bug-fixing-results} shows that bug fixes can reverse method rankings. 
On RoboTwin's \emph{Rotate QRCode}, the baseline shifts from
3 percentage points behind ProbeFlow to 6 ahead. On LIBERO-PRO's
\emph{Spatial-Task}, baseline SR rises from $0.4\%$ to $53.2\%$,
moving from last to first.

Extending the audit to VLABench shows effects even when the baseline
remains first. On \emph{Select Mahjong}, baseline SR gains 17 percentage points,
versus 6--10 for the accelerated methods. On \emph{Select Painting},
the baseline gains 13 points. Thus, bug fixes can also widen the baseline's lead
without changing the top-ranked method.

\begin{table}[!htb]
\centering
\vspace{-1em}
\addtolength{\belowcaptionskip}{2pt}
\caption{Original / revised SR (\%); blue/pink marks each row's worst/best values.
The listed LIBERO-PRO and LIBERO-Plus tasks are enhanced variants of a subset of LIBERO tasks.}
\label{tab:bug-fixing-results}
\small
\setlength{\tabcolsep}{3pt}
\renewcommand{\arraystretch}{1.14}
\newcommand{\srpair}[2]{\makebox[2.3em][r]{\ensuremath{#1}}\ensuremath{\,/\,}\makebox[2.3em][l]{\ensuremath{#2}}}
\begin{tabularx}{\linewidth}{@{}l@{\hspace{11pt}}>{\raggedright\arraybackslash}X@{\hspace{4pt}}cccc@{}}
\toprule
\textbf{Benchmark} & \textbf{Task / subset} & \textbf{Baseline}
& \textbf{DP-Cache-Fast}
& \textbf{DP-Cache-Slow} & \textbf{ProbeFlow} \\
\midrule
RoboTwin & \textit{place shoe} & \srpair{\evalworst{65}}{\evalbest{74}} & \srpair{\evalbest{68}}{69} & \srpair{66}{\evalworst{67}} & \srpair{\evalbest{68}}{69} \\
& \textit{rotate qrcode} & \srpair{56}{\evalbest{66}} & \srpair{\evalbest{60}}{62} & \srpair{\evalworst{48}}{\evalworst{57}} & \srpair{59}{60} \\
& $\cdots$ & & & & \\
& \textit{blocks ranking size} & \srpair{21}{\evalbest{27}} & \srpair{\evalbest{22}}{25} & \srpair{\evalworst{16}}{\evalworst{21}} & \srpair{19}{23} \\
& \textit{handover block} & \srpair{18}{\evalbest{29}} & \srpair{\evalworst{15}}{26} & \srpair{\evalbest{20}}{27} & \srpair{16}{\evalworst{25}} \\
\midrule
VLABench & \textit{select mahjong} & \srpair{\evalbest{18}}{\evalbest{35}} & \srpair{\evalworst{7}}{\evalworst{17}} & \srpair{14}{20} & \srpair{8}{\evalworst{17}} \\
& \textit{select painting} & \srpair{\evalbest{9}}{\evalbest{22}} & \srpair{8}{21} & \srpair{8}{19} & \srpair{\evalworst{6}}{\evalworst{14}} \\
\midrule
LIBERO-PRO & \textit{Object-Position} & \srpair{16.8}{\evalbest{18.6}} & \srpair{17.0}{17.6} & \srpair{\evalworst{15.8}}{\evalworst{17.0}} & \srpair{\evalbest{17.4}}{17.2} \\
 & \textit{Spatial-Position} & \srpair{46.4}{\evalbest{51.4}} & \srpair{\evalworst{45.6}}{\evalworst{50.6}} & \srpair{46.0}{51.0} & \srpair{\evalbest{47.8}}{\evalworst{50.6}} \\
& $\cdots$ & & & & \\
 & \textit{Object-Task} & \srpair{0.0}{\evalbest{10.6}} & \srpair{0.0}{\evalbest{10.6}} & \srpair{0.0}{\evalworst{10.0}} & \srpair{0.0}{\evalbest{10.6}} \\
 & \textit{Spatial-Task} & \srpair{\evalworst{0.4}}{\evalbest{53.2}} & \srpair{\evalworst{0.4}}{52.8} & \srpair{\evalbest{0.6}}{\evalworst{52.2}} & \srpair{\evalbest{0.6}}{\evalworst{52.2}} \\
\bottomrule
\end{tabularx}
\vspace{-1.2em}
\end{table}

\vspace{-0.8em}
\subsection{Results of Addressing Benchmark Limitations}
\label{sec:limitation-results}

\vspace{-0.6em}
\subsubsection{Permissive Checkers}
\label{sec:permissive-results}
\label{sec:threshold-results}

The \emph{Put Bottles Dustbin} case is detailed in
Section~\ref{sec:characterize-limitation}.
On VLABench's \emph{Select Drink} (Table~\ref{tab:mass-results}), the
original checker gives the baseline an SR of $35\%$ and ProbeFlow $19\%$.
After requiring controlled removal, baseline SR drops to $31\%$, whereas
ProbeFlow drops to $9\%$.
The baseline's lead thus widens from 16 to 22 percentage points.

For \emph{Scan Object} in Figure~\ref{fig:design-limitations}(b),
we vary the transverse tolerance $e$, keeping other checker conditions fixed.
Below $25$\,mm, methods achieve similar success rates; above
$25$\,mm, the baseline outperforms all accelerated methods,
especially at larger thresholds.
At $25$\,mm, the baseline trails ProbeFlow by 5 percentage points
($31\%$ vs.\ $36\%$); at $50$\,mm, it leads by 8 points
($58\%$ vs.\ $50\%$).

\vspace{-0.6em}
\subsubsection{Mass Sensitivity and Motion-Aware Scoring}
\vspace{-0.2em}
\label{sec:motion-results}
\label{sec:mass-results}

Table~\ref{tab:mass-results} presents selected tasks for mass adjustment and motion-aware scoring.
With the checkpoint and scoring rule held fixed, mass adjustment
reduces baseline SR on \emph{Place Fan} from $40\%$ to $38\%$,
while DP-Cache-Fast drops from $61\%$ to $33\%$
(Table~\ref{tab:mass-results}).
The baseline thus moves from 21 percentage points behind DP-Cache-Fast
to 5 points ahead.
Training with the corrected mass further raises baseline SR from
$38\%$ to $53\%$, outperforming DP-Cache-Fast ($33\%$-$45\%$) by 8 percentage points.

After motion-aware scoring, the baseline moves from 5 percentage points
behind ProbeFlow to 16.6 points ahead on \emph{Scan Object}.
On \emph{Turn Switch}, baseline SR remains at $25\%$, while DP-Cache-Slow
drops from $31\%$ to $24.8\%$.
These reversals highlight the importance of realistic physical
parameters and action-processing settings for reliable policy rankings.

\begin{table}[!htbp]
\centering
\vspace{-1.4em}
\caption{Original SR (\%) / revised SR or scores; blue/pink marks each row's worst/best values.}
\label{tab:mass-results}
\label{tab:motion-results}
\small
\setlength{\tabcolsep}{3pt}
\renewcommand{\arraystretch}{1.14}
\newcommand{\srchange}[2]{\makebox[1.6em][r]{\ensuremath{#1}}\ensuremath{\,/\,}\makebox[1.6em][l]{\ensuremath{#2}}}
\begin{tabularx}{\linewidth}{@{}l@{\hspace{16pt}}>{\raggedright\arraybackslash}X@{\hspace{4pt}}cccc@{}}
\toprule
\textbf{Benchmark} & \textbf{Task} & \textbf{Baseline}
& \textbf{DP-Cache-Fast} & \textbf{DP-Cache-Slow} & \textbf{ProbeFlow} \\
\midrule
\multicolumn{6}{@{}l}{\textbf{Permissive checkers}} \\
RoboTwin & \textit{put bottles dustbin} & \srchange{\evalworst{24}}{\evalbest{24}} & \srchange{26}{\evalworst{22}} & \srchange{\evalbest{28}}{\evalworst{22}} & \srchange{\evalbest{28}}{\evalworst{22}} \\
VLABench & \textit{select drink} & \srchange{\evalbest{35}}{\evalbest{31}} & \srchange{\evalworst{17}}{12} & \srchange{20}{17} & \srchange{19}{\evalworst{9}} \\
\midrule
\multicolumn{6}{@{}l}{\textbf{Mass adjustment}} \\
RoboTwin & \textit{beat block hammer} & \srchange{\evalworst{79}}{\evalbest{27}} & \srchange{\evalbest{87}}{\evalworst{19}} & \srchange{82}{\evalworst{19}} & \srchange{86}{22} \\
 & \textit{place fan} & \srchange{\evalworst{40}}{\evalbest{38}} & \srchange{\evalbest{61}}{\evalworst{33}} & \srchange{51}{\evalbest{38}} & \srchange{59}{34} \\
\addlinespace[3pt]
LIBERO & \textit{Mug Placement in Microwave} & \srchange{\evalworst{94}}{\evalbest{60}} & \srchange{\evalbest{98}}{56} & \srchange{\evalworst{94}}{52} & \srchange{\evalbest{98}}{\evalworst{50}} \\
 & \textit{Two-Mug Placement} & \srchange{\evalbest{98}}{\evalbest{88}} & \srchange{\evalworst{94}}{82} & \srchange{\evalworst{94}}{82} & \srchange{\evalworst{94}}{\evalworst{80}} \\
\midrule
\multicolumn{6}{@{}l}{\textbf{Motion-aware scoring}} \\
RoboTwin & \textit{scan object} & \srchange{\evalworst{31}}{\evalbest{31.0}} & \srchange{34}{20.0} & \srchange{35}{28.0} & \srchange{\evalbest{36}}{\evalworst{14.4}} \\
& \textit{turn switch} & \srchange{\evalworst{25}}{\evalbest{25.0}} & \srchange{\evalworst{25}}{\evalworst{12.8}} & \srchange{\evalbest{31}}{24.8} & \srchange{27}{13.2} \\
\bottomrule
\end{tabularx}
\vspace{-0.4em}
\end{table}

\vspace{-1.2em}
\section{Conclusion}
\label{sec:conclusion}
\vspace{-0.6em}

This work systematically investigates benchmark flaws to enable trustworthy comparisons of VLA acceleration methods.
Through a comprehensive evaluation across seven benchmarks, we identify 22 bugs and 4 limitations
involving instruction--checker mismatches, initialization defects, incomplete reproducibility,
success checks, physical parameters, and action-processing mechanisms.
We report these findings and contribute fixes to the community.
We will continue investigating flaws in emerging benchmarks and tracking the progress of submitted
pull requests to support more reliable evaluation of VLA acceleration methods.

\bibliography{paper}
\bibliographystyle{paper}

\clearpage
\appendix
\section{Hardware and Model Setup}
\label{app:hardware-models}

Our evaluation infrastructure uses NVIDIA A800 and RTX 4090 GPUs. We build on
public benchmark implementations and community-released $\pi_{0.5}$
checkpoints adapted to the corresponding tasks. Each method comparison uses
the same checkpoint and observation/action interface within a benchmark.
For RoboTwin, we use the $\pi_{0.5}$ checkpoint released by the Motus
team~\citep{motus}.\footnote{\url{https://huggingface.co/motus-robotics/pi0.5_robotwin2}}

Our real-robot experiments use the Mobile ALOHA platform~\citep{ALOHA}.

\begin{figure}[!htb]
\centering
\includegraphics[width=\linewidth]{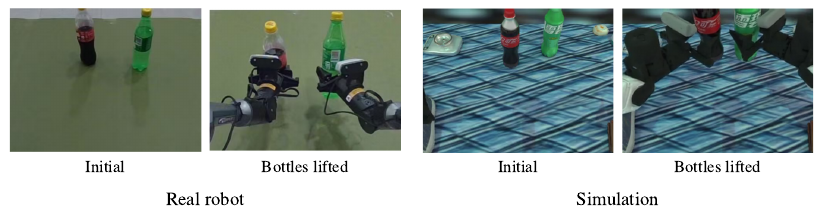}
\caption{\emph{Pick Dual Bottles} on the real robot (left pair) and in simulation (right pair).}
\label{fig:evaluation-examples}
\end{figure}

\subsection{Benchmark and Evaluation Framework Versions}
\label{app:benchmark-versions}

Table~\ref{tab:evaluation-commits} records the repository revisions used in
our evaluations. 

\begin{table}[htbp]
\centering
\caption{Repository revisions used for benchmark evaluation.}
\label{tab:evaluation-commits}
\small
\setlength{\tabcolsep}{5pt}
\begin{tabular}{@{}ll@{}}
\toprule
Repository & Commit SHA \\
\midrule
RoboTwin & \texttt{2eeec322d95799f537cbfe5f291a8220d965ccb8} \\
LIBERO & \texttt{8f1084e3132a39270c3a13ebe37270a43ece2a01} \\
LIBERO-PRO & \texttt{eafdb809426b13153aa1e4c42d6601844217dfec} \\
RoboCasa365 & \texttt{be22d659b02db8f6d7f3a3c3edc742934fdcbaae} \\
LIBERO-Plus & \texttt{4976dc30028e805ff8094b55501d532c48fec182} \\
VLABench & \texttt{cf588fe60c0c7282174fe979f5913170cfe69017} \\
RoboDojo & \texttt{ee67a1468510da7624a089164402359f2afc72c8} \\
\bottomrule
\end{tabular}
\end{table}

\subsection{Method Configurations and Comparison Protocol}
\label{app:method-configurations}

\paragraph{DP-Cache.}
Following the notation of \citet{dpcache}, $S$ denotes the cache interval:
one recomputation per $S$ steps, with cached outputs reused on the remaining
steps. DP-Cache-Fast and DP-Cache-Slow use $S=5$ and $S=2$, respectively.

\paragraph{ProbeFlow.}
We use linearity tolerance $\epsilon=0.008$, integration-step bounds
$N_{\min}=2$ and $N_{\max}=10$, step increment $\Delta N=2$, and lookahead
probe horizon $\Delta t_{\mathrm{probe}}=0.5$, following the settings reported
by \citet{probeflow}.

\paragraph{Comparison protocol.}
Repairs are compared within the same task cohort. When only the checker
changes, the comparison can use two scores from the same recorded trajectory;
prompt, scene, or physical parameter changes require a new rollout under the
corresponding condition.

\paragraph{Training details.}
For the 3,000-step \emph{place fan} fine-tuning run, we collect 100 successful
expert demonstrations with a 600 g fan and a 20 g target pad. We initialize from the existing Motus
$\pi_{0.5}$ checkpoint. Training uses one NVIDIA A800, batch size 4, a peak learning
rate of $2.5\times10^{-5}$, and 100 warm-up steps.

\subsection{Inference Latency}
\label{app:inference-latency}

We measure OpenPI PyTorch inference on one NVIDIA RTX 4090 with compilation
disabled, using the original RoboTwin $\pi_{0.5}$ checkpoint. Each configuration
receives 20 warm-up requests followed by 200 timed requests, with batch size 1
and action horizon 32. Inputs are fixed synthetic three-camera observations
($224\times224$), state, and a task prompt. 

\begin{table}[htbp]
\centering
\caption{Mean inference latency and speedup.}
\label{tab:inference-latency}
\small
\begin{tabular}{@{}lrr@{}}
\toprule
Configuration & Latency (ms) & Speedup \\
\midrule
Baseline & 313.73 & $1.00\times$ \\
DP-Cache-Fast & 122.61 & $2.56\times$ \\
DP-Cache-Slow & 194.12 & $1.62\times$ \\
ProbeFlow & 210.87 & $1.49\times$ \\
\bottomrule
\end{tabular}
\end{table}

\subsection{Measurement Details}
\label{app:measurement-details}
Output-action variation and executed joint jerk are reported separately,
using action indices and physical timestamps, respectively. Mass comparisons
use original and adjusted parameter settings while recording whether inertia
is also changed. A dash denotes an unfilled result, not zero success or a
completed experiment.

\subsection{Video-Recording Overhead}
\label{app:recording-overhead}
On device-local frozen \emph{place fan} cases, enabling additional video
recording increases mean episode latency from 86.42 s to 160.41 s on an
A800 (85\%) and from 24.22 s to 39.97 s on an RTX 4090 (65\%).
Policy-input rendering remains enabled in both conditions.

\section{Benchmark Bug Repairs and Design Limitations}
\label{app:extended-bug-results}

\begin{figure}[!htb]
\centering
\begin{minipage}[t]{0.47\linewidth}
\centering
\includegraphics[width=\linewidth]{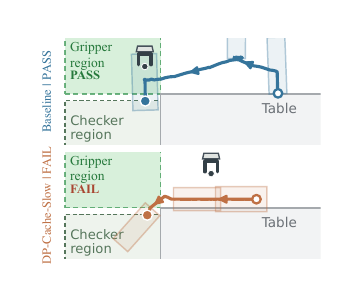}
\par\vspace{-0.08in}\small (a) Additional process check
\end{minipage}\hfill
\begin{minipage}[t]{0.47\linewidth}
\centering
\includegraphics[width=\linewidth]{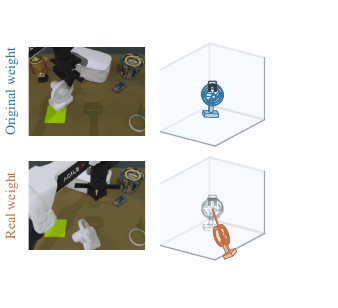}
\par\vspace{-0.08in}\small (b) Physical-parameter effect
\end{minipage}
\caption{(a) The gripper-presence check accepts Baseline and rejects DP-Cache-Slow. (b) An edge grasp holds the 10 g fan but fails at 600 g. Gray dashed poses and grippers are schematic references; pad mass and inertia also change.}
\label{fig:repair-limitation-examples}
\end{figure}

Table~\ref{tab:complete-bug-repairs} lists the 22 bugs identified in
Section~\ref{sec:benchmark-bugs} and their corresponding repairs, grouped
by category and subcategory. Figure~\ref{fig:repair-limitation-examples} illustrates a process-aware checker and the effect of physical-parameter changes.

\begingroup
\small
\setlength{\tabcolsep}{4pt}
\renewcommand{\arraystretch}{1.04}
\begin{longtable}{@{}p{.04\linewidth}p{.14\linewidth}>{\raggedright\arraybackslash}p{.44\linewidth}>{\raggedright\arraybackslash}p{\dimexpr.38\linewidth-24pt\relax}@{}}
\caption{Benchmark bugs and corresponding repairs.}\label{tab:complete-bug-repairs}\\
\toprule
No. & Benchmark & Bug & Repair \\
\midrule
\endfirsthead
\multicolumn{4}{l}{\tablename~\thetable\ (continued)}\\
\toprule
No. & Benchmark & Bug & Repair \\
\midrule
\endhead
\midrule
\multicolumn{4}{r}{Continued on next page}\\
\endfoot
\bottomrule
\endlastfoot
\multicolumn{4}{@{}l}{\textbf{Task consistency --- Semantic omission}}\\*
1 & RoboTwin & \textit{place shoe}: the checker and demonstrations restrict toe orientation, but the instruction only specifies mat placement. & Remove the unrequested orientation constraint; retain the mat-placement check. \\
2 & RoboTwin & \textit{rotate qrcode}: the checker adds table-height and gripper-release requirements absent from the instruction. & Remove the height and release constraints; retain the orientation check. \\
3 & RoboTwin & \textit{adjust bottle}: the checker restricts lateral position by initial side, although the prompt only requests grasping and lifting. & Accept either lateral region when no arm is specified; see Figure~\ref{fig:adjust-bottle-denoising} for arm-selection analysis. \\
4 & RoboTwin & \textit{blocks ranking size}: the instruction omits parts of the ordering rule used by the checker and demonstrations. & Explicitly state the required size order and placement in the instruction. \\
5 & RoboTwin & \textit{handover block}: grasp-only prompts are scored against a longer handover, placement, and release sequence. & Adapt the checker to the delivered prompt and check only its required operations. \\
6 & VLABench & \textit{select mahjong}: a pickup instruction is scored against an additional mat placement requirement. & remove the unrequested mat-placement condition. \\
\addlinespace[5pt]
\multicolumn{4}{@{}l}{\textbf{Task consistency --- Semantic conflict}}\\*
7 & RoboTwin & \textit{place bread skillet}: prompts specify the left arm, demonstrations use the right, and arm use is not checked. & Align the specified arm with demonstrations and add an arm-use check. \\
8 & RoboTwin & \textit{place bread basket}: dual-arm prompts conflict with single-arm demonstrations; only final positions are checked. & Check arm use against the prompt's single- or dual-arm requirement. \\
9 & VLABench & \textit{select painting}: prompts request grasping a painting, but demonstrations and scoring use a button press. & For grasping prompts, check a grasp of the target painting instead of a button press. \\
\addlinespace[5pt]
\multicolumn{4}{@{}l}{\textbf{Initialization --- Prompt pollution}}\\*
10 & LIBERO-PRO & Prompts come from stale filenames rather than updated BDDL instructions, causing conflicts such as stove on versus off. & Read the instruction from the current task's BDDL language field. Evaluation results are reported in Tables~\ref{tab:bug-fixing-results} and~\ref{tab:additional-bug-results}. \\
11 & LIBERO-Plus & BDDL language fields contain LLM-rewriting preambles. & Correct the task language in the affected BDDL files. \\
12 & LIBERO-Plus & Filename-derived prompts include camera views, initial-state indices, or other configuration metadata. & Read task language from BDDL to exclude filename metadata from policy inputs. Evaluation results are reported in Table~\ref{tab:additional-bug-results}. \\
\addlinespace[5pt]
\multicolumn{4}{@{}l}{\textbf{Initialization --- Invalid initial states}}\\*
13 & LIBERO-PRO & Perturbed scenes load incompatible old states, leaving object positions inconsistent with the new scene. & Align generation/loading directories, load matching states, and check initial spatial relations. \\
14 & RoboCasa & \textit{ArrangeDrinkware}: identical source and destination counters make the task successful at initialization. & Resample initially successful instances. After repair, $\pi_{0.5}$ SR drops from $20\%$ to $0\%$. \\
15 & RoboDojo & \textit{organize table}: the prompt requests drawer placement of objects absent from the initial scene. & Add the objects required for drawer placement and the corresponding success checks. \\
\addlinespace[5pt]
\multicolumn{4}{@{}l}{\textbf{Reproducibility --- Incomplete evaluation configuration}}\\*
16 & RoboTwin & Instruction sampling is not fully seeded, so identical episode seeds can yield different prompts. & Bind instruction-generation and selection randomness to the episode seed. \\
17 & VLABench & The environment loader omits the random state required by the independent scene generator. & Pass a seed-controlled random state to the scene generator. \\
18 & LIBERO-Plus & Non-default task orders are sampled before setting the command-line seed. & Set the seed before generating the task order. \\
19 & LIBERO & Unpinned MuJoCo versions produce different rendered policy inputs. & Pin MuJoCo to 3.3.2 instead of 3.8.1; $\pi_{0.5}$ SR on \textit{pick up bowl} rises from $62\%$ to $88\%$. \\
\addlinespace[5pt]
\multicolumn{4}{@{}l}{\textbf{Reproducibility --- History-dependent scene construction}}\\*
20 & RoboTwin & In-place changes to clutter bounds accumulate across episodes, making scenes depend on initialization history. & Initialize from a fresh copy of the original bounds to prevent accumulated offsets. \\
21 & RoboCasa & History-dependent resets can produce different scenes for the same seed and trial index. & Add a reset interface with explicit per-episode seeding. \\
\addlinespace[5pt]
\multicolumn{4}{@{}l}{\textbf{Reproducibility --- Incomplete state restoration}}\\*
22 & RoboCasa & Restored scenes can be scored against targets from a newly sampled scene, reversing a saved success label. & Save and restore checker targets and other task state alongside the simulation state. \\
\end{longtable}
\endgroup

\paragraph{Process-aware dustbin checker.}
For \emph{put bottles dustbin}, we measured the time for a bottle to fall
into the bin from positions near the table edge, at heights ranging from
tabletop level to 30 cm above it, and obtained a 250 ms interval.
We therefore augment the original bottle-position checks with a per-bottle
process check: during the 250 ms preceding entry into the checker region,
the gripper held the bottle  must have appeared in the space directly above that
region. Applying this condition to each bottle adds evidence of delivery
by the robot rather than accepting position alone.
Figure~\ref{fig:repair-limitation-examples}(a) illustrates one bottle:
Baseline satisfies the additional check, whereas DP-Cache-Slow does not.

\paragraph{Acceptance-threshold sensitivity.}
In \emph{scan object}, $d$ limits axial separation and $e$ bounds transverse
alignment error component-wise. We fix $d=70$\,mm (default) and, among states
satisfying all other checks, record each episode's minimum required tolerance
$e_{\min}$. Figure~\ref{fig:design-limitations}(b) plots the fraction with
$e_{\min}<e$ as $e$ varies. 

\paragraph{Mass-sensitive grasping.}
Figure~\ref{fig:repair-limitation-examples}(b) illustrates the effect of
correcting the fan mass in \emph{place fan}. A grasp near the fan's edge
can succeed in simulation with the original 10 g mass. After the fan mass
is corrected to 600 g, the illustrated grasp no longer retains the fan:
it slips and falls, and the task fails.

\FloatBarrier

\paragraph{Motion-aware ranking.}
\label{app:motion-ranking}
For each paired episode, we rank all four methods by the mean absolute
action jerk magnitude of their policy-output actions over time and the
12 arm joints, excluding grippers. Ranks 1--4 receive
1.0, 0.8, 0.6, and 0.4 points (ties share the average); failures participate
in ranking but receive zero points. We report $100$ times the mean episode
score, alongside SR, in Tables~\ref{tab:mass-results}
and~\ref{tab:additional-mass-motion}.

\begin{figure}[!htb]
\centering
\includegraphics[width=\linewidth]{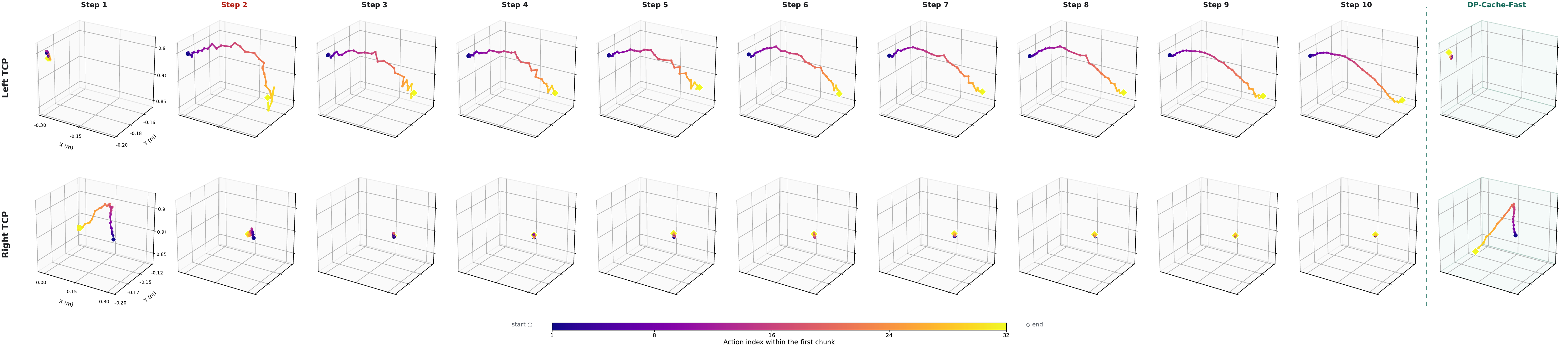}
\caption{Step-by-step flow-matching denoising on \emph{adjust bottle}.
Top/bottom rows show left/right TCP paths for the first 32-action chunk;
color indicates action order within the chunk. The rightmost column
shows the accelerated comparison. From Step 2, motion remains concentrated
in the left arm and becomes smoother over subsequent denoising steps.}
\label{fig:adjust-bottle-denoising}
\end{figure}

\paragraph{Task consistency: a closer look at arm selection during denoising.}
We further investigate the task-consistency case study through a more
detailed analysis of the denoising process.
Figure~\ref{fig:adjust-bottle-denoising} examines the first action chunk of
an \emph{adjust bottle} case across ten flow-matching denoising steps.
TCP denotes the \emph{tool center point}, the reference point at the gripper
end effector. From Step 2 onward, the dominant predicted motion
consistently shifts to the left arm, while the right arm remains nearly
stationary. As denoising proceeds, the predicted left-arm path becomes
visually smoother overall. This case illustrates how denoising can change
arm selection and refine motion, motivating the prompt-aligned lateral
acceptance rule described in entry 3 of Table~\ref{tab:complete-bug-repairs}.

\section{Additional Success-Rate Comparisons and Visualizations}
\label{app:additional-results}

\begin{figure}[p]
\centering
\setlength{\abovecaptionskip}{5pt}
\includegraphics[width=0.95\linewidth]{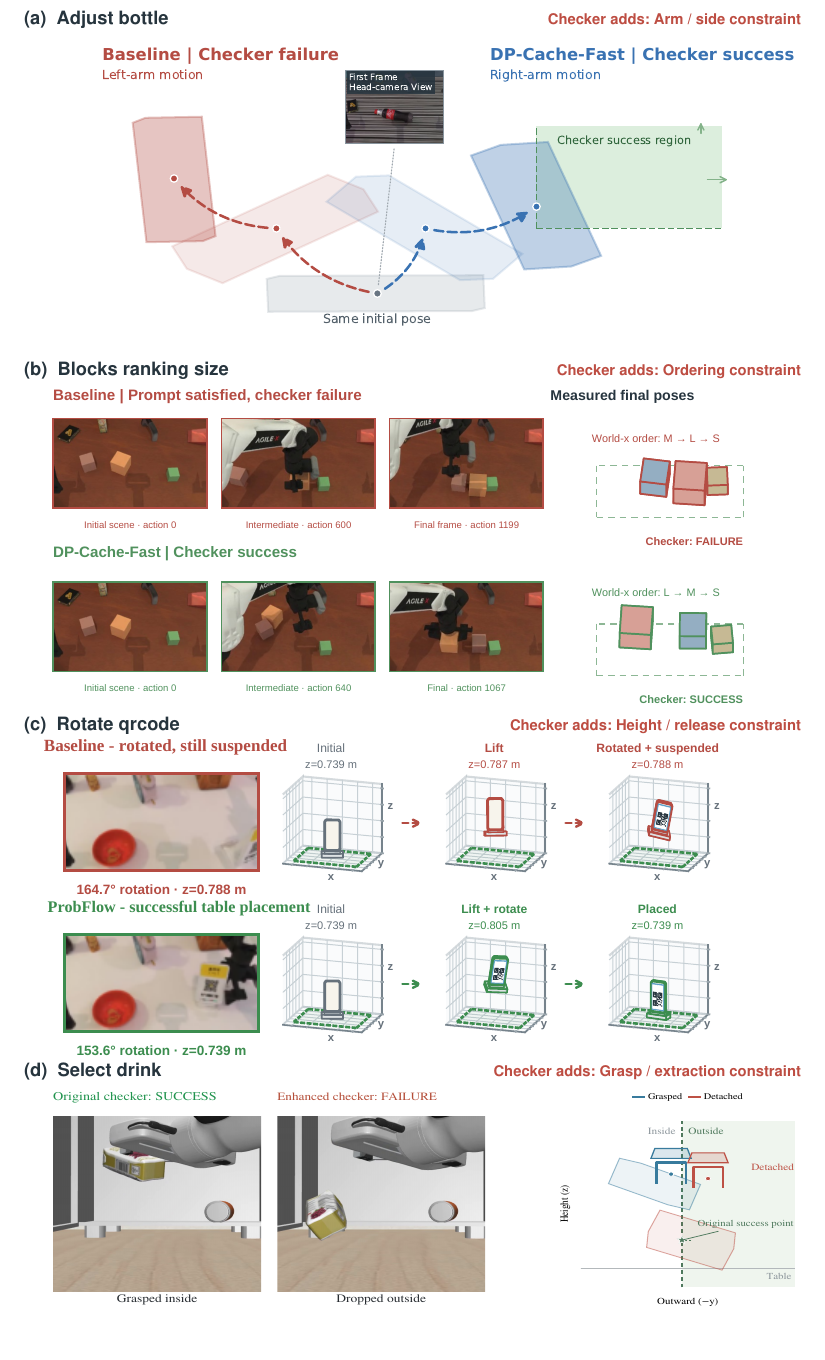}\vspace{-24pt}
\caption{Visual examples of benchmark issues. Bug details and repairs are provided in Table~\ref{tab:complete-bug-repairs}.}
\label{fig:robotwin-task-consistency-cases}
\end{figure}

Table~\ref{tab:additional-bug-results} supplements the bug-fix comparisons
in Table~\ref{tab:bug-fixing-results}. Figure~\ref{fig:robotwin-task-consistency-cases} provides visual examples of instruction--checker mismatches and permissive success checks. Table~\ref{tab:additional-mass-motion}
adds mass-adjustment and motion-aware scoring results omitted from
Table~\ref{tab:mass-results}. Entries report before / after SR (\%), except motion-aware rows, which report SR (\%) / score (0--100).

\begin{table}[!htb]
\centering
\caption{Additional bug-fix comparisons: original / repaired SR (\%); blue/pink marks each row's worst/best values separately before and after repair.}
\label{tab:additional-bug-results}
\providecommand{\srpair}[2]{\makebox[2.3em][r]{\ensuremath{#1}}\ensuremath{\,/\,}\makebox[2.3em][l]{\ensuremath{#2}}}
\small
\setlength{\tabcolsep}{3pt}
\renewcommand{\arraystretch}{1.04}
\begin{tabularx}{\linewidth}{@{}l>{\raggedright\arraybackslash}Xcccc@{}}
\toprule
Benchmark & Task & Baseline & DP-Cache-Fast & DP-Cache-Slow & ProbeFlow \\
\midrule
\multicolumn{6}{@{}l}{\textbf{Task consistency}} \\
RoboTwin & \textit{adjust bottle} & \srpair{\evalworst{92}}{\evalbest{96}} & \srpair{94}{\evalworst{94}} & \srpair{\evalbest{95}}{95} & \srpair{94}{\evalworst{94}} \\
& \textit{place bread skillet} & \srpair{42}{\evalbest{35}} & \srpair{\evalbest{47}}{32} & \srpair{\evalworst{39}}{\evalworst{29}} & \srpair{46}{32} \\
& \textit{place bread basket} & \srpair{\evalworst{55}}{\evalbest{54}} & \srpair{\evalbest{59}}{53} & \srpair{\evalworst{55}}{\evalworst{52}} & \srpair{58}{\evalworst{52}} \\
\midrule
\multicolumn{6}{@{}l}{\textbf{Initialization}} \\
LIBERO-PRO & \textit{Goal-Language} & \srpair{97.2}{96.4} & \srpair{\evalbest{98.4}}{96.6} & \srpair{\evalworst{96.6}}{\evalworst{95.8}} & \srpair{\evalbest{98.4}}{\evalbest{96.8}} \\
 & \textit{Object-Language} & \srpair{\evalworst{97.8}}{\evalworst{98.8}} & \srpair{\evalbest{98.8}}{\evalbest{99.8}} & \srpair{98.4}{99.2} & \srpair{\evalbest{98.8}}{\evalbest{99.8}} \\
 & \textit{Spatial-Object} & \srpair{\evalworst{96.8}}{96.2} & \srpair{\evalbest{98.2}}{\evalbest{98.4}} & \srpair{97.0}{\evalworst{95.8}} & \srpair{98.0}{97.2} \\
 & \textit{Goal-Position} & \srpair{34.0}{\evalbest{34.0}} & \srpair{\evalworst{31.6}}{\evalworst{31.0}} & \srpair{\evalbest{34.4}}{33.6} & \srpair{32.8}{32.4} \\
 & \textit{Goal-Task} & \srpair{\evalworst{8.6}}{\evalworst{20.0}} & \srpair{\evalbest{9.0}}{\evalbest{22.2}} & \srpair{\evalworst{8.6}}{21.2} & \srpair{\evalbest{9.0}}{20.2} \\
\addlinespace[2pt]
LIBERO-Plus & \textit{object-camera} & \srpair{\evalworst{84.34}}{\evalbest{92.42}} & \srpair{88.38}{\evalbest{92.42}} & \srpair{85.35}{\evalworst{91.92}} & \srpair{\evalbest{90.15}}{92.17} \\
\bottomrule
\end{tabularx}
\end{table}

\begin{table}[!htb]
\centering
\caption{Additional results: before / after SR (\%) for mass adjustment; SR (\%) / score (0--100) for motion-aware scoring. Blue/pink marks each row's worst/best values separately for each condition, including ties.}
\label{tab:additional-mass-motion}
\providecommand{\srpair}[2]{\makebox[2.3em][r]{\ensuremath{#1}}\ensuremath{\,/\,}\makebox[2.3em][l]{\ensuremath{#2}}}
\small
\setlength{\tabcolsep}{3pt}
\renewcommand{\arraystretch}{1.14}
\begin{tabularx}{\linewidth}{@{}l>{\raggedright\arraybackslash}Xcccc@{}}
\toprule
Benchmark & Task & Baseline & DP-Cache-Fast & DP-Cache-Slow & ProbeFlow \\
\midrule
\multicolumn{6}{@{}l}{\textbf{Mass adjustment}} \\
RoboTwin & \textit{place a2b right} & \srpair{71}{\evalbest{68}} & \srpair{74}{\evalworst{59}} & \srpair{\evalbest{78}}{62} & \srpair{\evalworst{68}}{62} \\
 & \textit{stack blocks three} & \srpair{26}{\evalbest{34}} & \srpair{26}{31} & \srpair{\evalworst{18}}{\evalworst{26}} & \srpair{\evalbest{28}}{31} \\
\addlinespace[3pt]
LIBERO & \textit{Plate Pushing} & \srpair{\evalbest{100}}{86} & \srpair{\evalbest{100}}{\evalbest{90}} & \srpair{\evalbest{100}}{\evalworst{80}} & \srpair{\evalbest{100}}{88} \\
\midrule
\multicolumn{6}{@{}l}{\textbf{Motion-aware scoring}} \\
RoboTwin & \textit{pick diverse bottles} & \srpair{37}{\evalbest{37.0}} & \srpair{\evalworst{36}}{18.2} & \srpair{38}{30.4} & \srpair{\evalbest{41}}{\evalworst{17.8}} \\
 & \textit{pick dual bottles} & \srpair{49}{\evalbest{48.6}} & \srpair{49}{25.2} & \srpair{\evalworst{46}}{36.8} & \srpair{\evalbest{50}}{\evalworst{21.6}} \\
\bottomrule
\end{tabularx}
\end{table}

Table~\ref{tab:mass-settings} lists the RoboTwin and LIBERO mass changes in
grams. In LIBERO, the yellow-and-white mug in \emph{Mug Placement in Microwave}
(including door closure) changes from 30.91 to 300 g ($9.7\times$); the plate
in \emph{Plate Pushing} from 11.52 to 150 g; and the white porcelain mug in
\emph{Two-Mug Placement} from 16.46 to 250 g. The latter task's SR comparison
appears in Table~\ref{tab:mass-results}.

\clearpage
\begin{table}[!htb]
\centering
\caption{Object masses used in the RoboTwin and LIBERO mass-adjustment comparisons (g).}
\label{tab:mass-settings}
\small
\setlength{\tabcolsep}{6pt}
\renewcommand{\arraystretch}{1.04}
\begin{tabularx}{\linewidth}{@{}>{\raggedright\arraybackslash}Xlrr@{}}
\toprule
Task & Object & Original (g) & Adjusted (g) \\
\midrule
\multicolumn{4}{@{}l}{\textbf{RoboTwin}} \\
\textit{beat block hammer} & Hammer & 1 & 600 \\
\textit{place a2b right} & Bell & 50 & 150 \\
\textit{stack blocks three} & Block & 10 & 50 \\
\textit{place fan} & Fan & 10 & 600 \\
\midrule
\multicolumn{4}{@{}l}{\textbf{LIBERO}} \\
\textit{Mug Placement in Microwave} & Yellow-and-white mug & 30.91 & 300 \\
\textit{Plate Pushing} & Plate & 11.52 & 150 \\
\textit{Two-Mug Placement} & White porcelain mug & 16.46 & 250 \\
\bottomrule
\end{tabularx}
\end{table}

\section{Practical Guidelines for Assessing Acceleration Gains}
\label{app:practical-guidelines}
The repairs above motivate four practical recommendations for assessing
whether task-specific gains under acceleration reflect valid task completion
under fair comparison conditions.

\newcommand{\appendixguidelinebox}[1]{%
  \par\addvspace{0.35\baselineskip}\noindent
  \begingroup
  \setlength{\fboxsep}{3pt}%
  \setlength{\fboxrule}{0.3pt}%
  \fcolorbox{black!30}{black!5}{%
    \parbox{\dimexpr\linewidth-2\fboxsep-2\fboxrule\relax}{\bfseries #1}}%
  \endgroup\par\nobreak\vspace{0.25\baselineskip}\noindent\ignorespaces
}

\appendixguidelinebox{Advice \#1. Make robots follow human instructions.}
Policies should follow instructions rather than rely on memorized scenes.
Align expert demonstrations and success checks with the requested actions
and outcomes. State necessary constraints explicitly and remove checks
that impose unstated requirements or reward unintended behavior. Document
repairs to existing benchmarks and enforce this consistency when building
new ones.

\appendixguidelinebox{Advice \#2. Validate the task the policy actually receives.}
Inspect the delivered prompt and the fully initialized scene, since loading
and perturbation mechanisms can change the task presented to the policy.
Check for incorrect or irrelevant text in instructions, mismatches between the scene and its
initial state, and unavailable operations. Verify that initial spatial
relations match the task description and that success is not already
satisfied before execution.

\appendixguidelinebox{Advice \#3. Make baseline and accelerated runs use matched task instances.}
A shared seed does not establish that two runs instantiate the same task.
Control all random sources and save the information needed to restore the
scene and its scoring targets. Verify that baseline and accelerated runs
receive matching instructions, initial states, and initial model inputs,
with identical success criteria. Test state restoration and action replay
separately to establish which aspects of an episode can be reproduced.

\appendixguidelinebox{Advice \#4. Beyond success rate: motion quality and completion time.}
Report motion jitter and end-to-end completion time alongside success rate.
Distinguish policy-output variation from executed-motion jerk, using a
consistent time base for each. Include both inference and robot motion
in completion time, and report failures and timeouts so early termination
is not mistaken for efficiency.

The bugs and design limitations uncovered by our analysis of anomalous
gains provide diagnostic leads for other benchmarks and tasks: examine
whether similar mechanisms occur and account for apparent improvements
under acceleration. Where such issues are found, compare the baseline and
accelerated policies on matched task instances before and after repairing
the bugs or strengthening the evaluation criteria to address the
limitations. Use trajectory evidence to examine both failures that become
successes and successes that become failures across task types, object
layouts, and operation stages. Gains that persist under these checks
provide stronger evidence that an acceleration method is well suited to
the task, whereas gains that disappear or reverse suggest that the apparent
task affinity may stem from benchmark flaws.

\clearpage
\clearpage

\end{document}